%% file: 0508.tex
\documentclass[runningheads]{llncs}
\usepackage{graphicx}
\usepackage{amsmath,amssymb} 
\usepackage{color}
\usepackage{microtype}
\usepackage{hyperref}
\usepackage{booktabs}

\begin{document}
\title{Recycle-GAN: Unsupervised Video Retargeting} 

\titlerunning{Recycle-GAN: Unsupervised Video Retargeting}
%
\author{Aayush Bansal\inst{1}\and
Shugao Ma\inst{2} \and
Deva Ramanan\inst{1} \and
Yaser Sheikh\inst{1,2}}
\authorrunning{A. Bansal, S. Ma, D. Ramanan, Y. Sheikh}
\institute{$^{1}$Carnegie Mellon University \quad $^{2}$Facebook Reality Lab, Pittsburgh\\
	\url{http://www.cs.cmu.edu/~aayushb/Recycle-GAN/}}
\maketitle            

\begin{abstract}
We introduce a data-driven approach for unsupervised video retargeting that translates content from one domain to another while preserving the style native to a domain, i.e., if contents of John Oliver's speech were to be transferred to Stephen Colbert, then the generated content/speech should be in  Stephen Colbert's style. Our approach combines both spatial and temporal information along with adversarial losses for content translation and style preservation. In this work, we first study the advantages of using spatiotemporal constraints over spatial constraints for effective retargeting.  We then demonstrate the proposed approach for the problems where information in both space and time matters such as face-to-face translation, flower-to-flower, wind and cloud synthesis, sunrise and sunset.

\end{abstract}

\input{introduction}

\input{related_work}

\input{approach}

\input{experiments}

\input{discussion}

\bibliographystyle{splncs04}
\bibliography{references}

\end{document}

%% file: introduction.tex
\section{Introduction}
\label{into}

\begin{figure*}[t]
  \includegraphics[width=\textwidth]{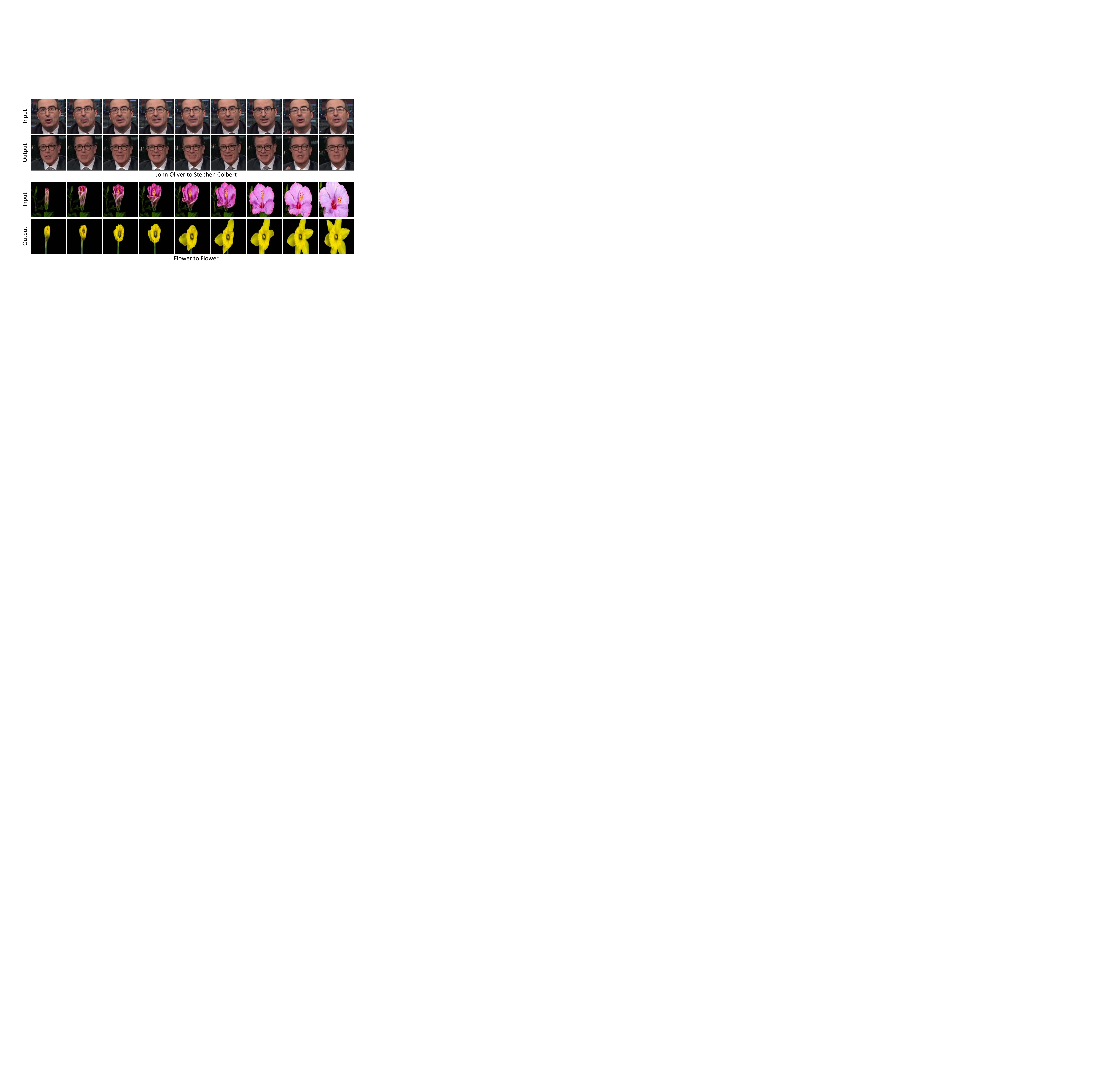}
   \caption{Our approach for video retargeting used for faces and flowers. The top row shows translation from John Oliver to Stephen Colbert. The bottom row shows how a synthesized flower follows the blooming process with the input flower. The corresponding videos are available on the project webpage.}
  \label{fig:teaser}
\end{figure*}

We present an unsupervised data-driven approach for video retargeting that enables the transfer of sequential content from one domain to another while preserving the style of the target domain. Such a content translation and style preservation task has numerous applications including human motion and face translation from one person to other, teaching robots from human demonstration, or converting black-and-white videos to color. This work also finds application in creating visual content that is hard to capture or label in real world settings, e.g., aligning human motion and facial data of two individuals for virtual reality, or labeling night data for a self-driving car. Above all, the notion of content translation and style preservation transcends pixel-to-pixel operation, into a more semantic and abstract human understandable concepts.

Current approaches for retargeting can be broadly classified into three categories. The first set is specifically designed for domains such as human faces ~\cite{Cao:2014:DDE,Thies:2015:RET,Thies_2016_CVPR}. While these approaches work well when faces are fully visible, they fail when applied to occluded faces (virtual reality) and  lack generalization to other domains. The work on paired image-to-image translation~\cite{pix2pix2016} attempts to generalize across domain but requires manual supervision for labeling and alignment. This requirement makes it hard for the use of such approaches as manual alignment or labeling is not possible in many domains. The third category of work attempts unsupervised and unpaired image translation~\cite{pmlr-v70-kim17a,ZhuPIE17}. They enforce a cyclic consistency~\cite{zhou2016learning} on unpaired 2D images and learn a transformation from one domain to another. However, unpaired 2D images alone are not sufficient for video retargeting. Firstly, it is not able to pose sufficient constraints on optimization and often leads to bad local minima or a perceptual mode collapse making it hard to generate the required output in the target domain. Secondly, the use of the spatial information alone in 2D images makes it hard to learn the \emph{style} of a particular domain as stylistic information requires temporal knowledge as well.

In this work, we make two specific observations: (i) the use of temporal information provides more constraints to the optimization for transforming one domain to other and helps in reaching a better local minima; (ii) the combined influence of spatial and temporal constraints helps in learning the style characteristic of an identity in a given domain. Importantly, temporal information is freely available in videos (available in abundance on web) and therefore no manual supervision is required. Figure~\ref{fig:teaser} shows an example each of translation for human faces, and flowers. Without any manual supervision and domain-specific knowledge, our approach learns this \emph{retargeting} from one domain to the other using publicly available video data on the web from both domains. 

\noindent\textbf{Our contributions} : We introduce a new approach that incorporates spatiotemporal cues with conditional generative adversarial networks~\cite{GoodfellowPMXWOCB14} for video retargeting. We demonstrate the advantages of spatiotemporal constraints over spatial constraints for image-to-labels,  and labels-to-image in diverse environmental settings. We then present the proposed approach in learning better association between two domains, and its importance for self-supervised content alignment of the visual data. Inspired by the ever-existing nature of space-time, we qualitatively demonstrate the effectiveness of our approach for various natural processes such as face-to-face translation, flower-to-flower, synthesizing clouds and winds, aligning sunrise and sunset.

%% file: related_work.tex
\section{Related Work}
\label{related}

\begin{figure}[t]
\centering
\includegraphics[width=1.0\linewidth]{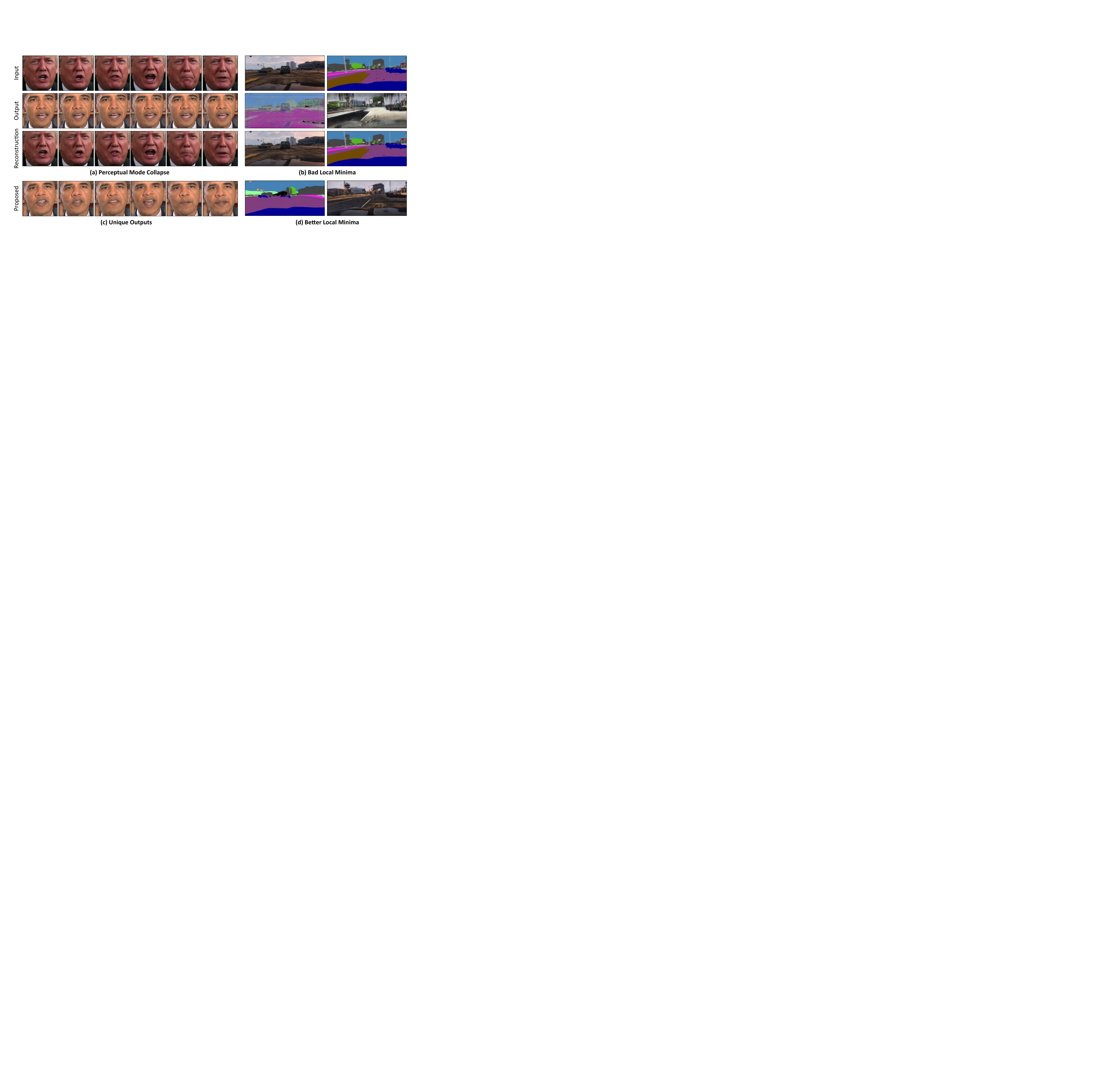}
\caption{\textbf{Spatial cycle consistency is not sufficient:}  We show two examples illustrating why spatial cycle consistency alone is not sufficient for the optimization. (a) shows an example of \textit{perceptual} mode-collapse while using Cycle-GAN~\cite{ZhuPIE17} for Donald Trump to Barack Obama. First row shows the input of Donald Trump, and second row shows the output generated. The third row shows the output of reconstruction that takes the second row as input. The second row looks similar despite different inputs; and the third row shows output similar to first row. On a very close observation, we found that a few pixels in second row were different (but not perceptually significant) and that was sufficient to get the different reconstruction; (b) shows another example for image-to-labels and labels-to-image. While the generator is not able to generate the required output for the given input in both the cases, it is still able to perfectly reconstruct the input. Both the examples suggest that the spatial cyclic loss is not sufficient to ensure the required output in another domain because the overall optimization is focussed on reconstructing the input. However as shown in (c) and (d) , \textbf{we get better outputs with our approach combining the spatial and temporal constraints}. Videos for face comparison are available on project webpage.}
\label{fig:problems}
\end{figure}

A variety of work dealing with image-to-image translation~\cite{Gatys_2016_CVPR,Hertzmann:2001,pix2pix2016,shrivastava2016learning,ZhuPIE17} and style translation~\cite{Brand:2000:SM,Freeman1997,Hsu:2005:STH} exists.  In fact a large body of work in computer vision and computer graphics is about an image-to-image operation.  While the primary efforts were on inferencing semantic~\cite{Long15}, geometric~\cite{Bansal16,Eigen15}, or low-level cues~\cite{Xie15}, there is a renewed interest in synthesizing images using data-driven approaches by the introduction of generative adversarial networks~\cite{GoodfellowPMXWOCB14}. This formulation has been used to generate images from cues such as a low-resolution image~\cite{DentonCSF15,LedigTHCATTWS16}, class labels~\cite{pix2pix2016}, and various other input priors~\cite{HuangLPHB16,RadfordMC15,ZhangXLZHWM16}. These approaches, however, require an input-output pair to train a model. While it is feasible to label data for a few image-to-image operations, there are numerous tasks for which it is non-trivial to generate input-output pairs for training supervision. Recently, Zhu et al.~\cite{ZhuPIE17} propose to use the  cycle-consistency constraint~\cite{zhou2016learning} in adversarial learning framework to deal with this problem of unpaired data, and demonstrate effective results for various tasks. Cycle-consistency~\cite{pmlr-v70-kim17a,ZhuPIE17} enables many image-to-image translation tasks without any expensive manual labeling. Similar ideas have also found application in learning depth cues in an unsupervised manner~\cite{GodardAB16}, machine translation~\cite{XiaHQWYLM16}, shape correspondences~\cite{Huang:2013}, point-wise correspondences~\cite{zhou2016learning,Zhou2015FlowWebJI}, or domain adaptation~\cite{Hoffman17CycADA}. 

The variants of Cycle-GAN~\cite{ZhuPIE17} have been applied to various temporal domains~\cite{GodardAB16,Hoffman17CycADA}. However, they consider only the spatial information in 2D images, and ignore the temporal information for optimization. We observe two major limitations: (1). \textbf{Perceptual Mode Collapse:} there are no guarantees that cycle consistency would produce perceptually unique data to the inputs. In Figure~\ref{fig:problems}, we show the outputs of a model trained for Donald Trump to Barack Obama, and an example for image-to-labels and labels-to-image. We find that for different inputs of Donald Trump, we get perceptually similar outputs of Barack Obama. We observe that these outputs have some unique encoding that enables them to reconstruct images similar to the inputs. We see similar behavior for image-to-labels and labels-to-image in Figure~\ref{fig:problems}-$(b)$; (2). \textbf{Tied Spatially to Input: } Due to the reconstruction loss on the input itself, the optimization is forced to learn a solution that is closely tied to the input. While this is reasonable for the problems where only spatial transformation matters (such as horse-to-zebra, apples-to-oranges, or paintings etc.), it is important for the problems where temporal and stylistic information is required for synthesis (prominently face-to-face translation). In this work, we propose a new formulation that utilizes both spatial and temporal constraints along with the adversarial loss to overcome these two problems. In Figure~\ref{fig:problems}-$(c,d)$, we show the outputs generated using proposed formulation that overcomes the above mentioned problems. We posit that this is due to more constraints available for an under-constrained optimization.

The use of GANs~\cite{GoodfellowPMXWOCB14} and variational auto-encoder~\cite{kingma2013auto} have also found a way for synthesizing videos and temporal information. Walker et al.~\cite{vae_eccv2016} use temporal information to predict future trajectories from a single image. Recent work~\cite{JHeVidGen2018,villegas2017learning,pos_iccv2017} used temporal models to predict long term future poses from a single 2D image. MoCoGAN~\cite{tulyakov2017mocogan} decomposes motion and content to control video generation. Similarly, Temporal GAN~\cite{TGAN2017} employs a temporal generator and an image generator that generates a set of latent variables and image sequences respectively. While relevant, this prior work focuses on predicting future intent from single images at test time or generating videos from a random noise. Concurrently, MoCoGAN~\cite{tulyakov2017mocogan} shows examples  of image-to-video translation using their formulation. Different from these methods, our focus is on a general video-to-video translation where the input video can control the output in a spirit similar to image-to-image translation. To this end, we can generate high-resolution videos of arbitrary length using our approach whereas prior work~\cite{TGAN2017,tulyakov2017mocogan} generate only 16 frames of $64 \times 64$. 

\begin{figure*}[t]
\centering
\includegraphics[width=1\linewidth]{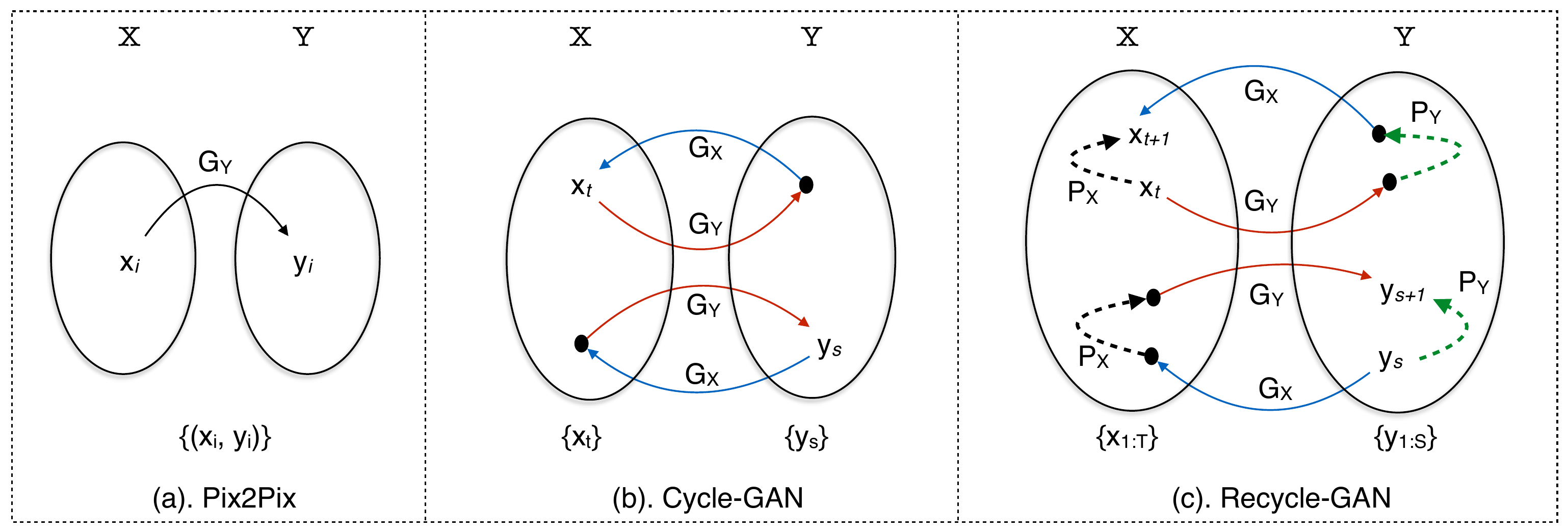}
\caption{We contrast our work with two prominent directions in image-to-image translation. (a) \textbf{Pix2Pix}~\cite{pix2pix2016}:  Paired data is available. A simple function (Eq.~\ref{eq:reg}) can be learnt via regression to map $X \rightarrow Y$. (b) \textbf{Cycle-GAN}~\cite{ZhuPIE17}:  The data is not paired in this setting. Zhu et al.~\cite{ZhuPIE17} proposed to use cycle-consistency loss (Eq.~\ref{eq:cycle}) to deal with the problem of unpaired data. (c) \textbf{Recycle-GAN}: The approaches so far have considered independent 2D images only. Suppose we have access to unpaired but {\em ordered streams} $(x_1,x_2,\ldots,x_t,\ldots)$ and $(y_1,y_2\ldots, y_s,\ldots)$. We present an approach that combines spatiotemporal constraints (Eq.~\ref{eq:cycle2}). See Section~\ref{approach} for more details.}
\label{fig:overview}
\end{figure*}

\noindent\textbf{Spatial \& Temporal Constraints} : Spatial and temporal information is known to be an integral sensory component that guides human action~\cite{gibson1979ecological}. There exists a wide literature utilizing these two constraints for various computer vision tasks such as learning better object detectors~\cite{MisraSSL15}, action recognition~\cite{Girdhar_17a_ActionVLAD} etc. In this work, we take a first step to exploit spatiotemporal constraints for video retargeting and unpaired image-to-image translation. 

\noindent\textbf{Learning Association: } Much of computer vision is about learning association, be it learning high-level image classification~\cite{Russakovsky15}, object relationships~\cite{malisiewicz-nips09}, or point-wise correspondences~\cite{pixelnn,KanazawaJC16,Liu:2011,LongNIPS14}. However, there has been relatively little work on learning association for aligning the content of different videos. In this work, we use our model trained with spatiotemporal constraints to align the semantical content of two videos in a self-supervised manner, and do automatic alignment of the visual data without any additional supervision.

%% file: approach.tex
\section{Method}
\label{approach}

Assume we wish to learn a mapping $G_Y : X \rightarrow Y$. The classic approach tunes $G_Y$ to minimize reconstruction error on paired data samples $\{(x_i,y_i)\}$ where $x_i \in X$ and $y_i \in Y$:  
\begin{align}
  \min_{G_Y} \sum_i ||y_i - G_Y(x_i)||^2 .
\label{eq:reg}
\end{align}

\noindent\textbf{Adversarial loss:} Recent work~\cite{pix2pix2016,GoodfellowPMXWOCB14} has shown that one can improve the learned mapping by tuning it with a discriminator $D_Y$ that is adversarially trained to distinguish between real samples of $y$ from generated samples $G_Y(x)$:
\begin{align}
\min_{G_Y} \max_{D_Y} L_{g}(G_Y,D_Y) =  \sum_s \log D_Y(y_s) + \sum_t \log (1 - D_Y(G_Y(x_t))) ,
\label{eq:adv}
\end{align}
Importantly, we use a formulation that does {\em not} require paired data and only requires access to individual samples $\{x_t\}$ and $\{y_s\}$, where different subscripts are used to emphasize the lack of pairing.

\noindent\textbf{Cycle loss: } Zhu et al.~\cite{ZhuPIE17} use cycle consistency~\cite{zhou2016learning} to define a reconstruction loss when the pairs are not available. Popularly known as Cycle-GAN (Fig.~\ref{fig:overview}-b), the objective can be written as:
\begin{align}
L_c(G_{X}, G_{Y}) =  \sum_t ||x_t - G_{X}(G_{Y}(x_t))||^2 .
\label{eq:cycle}
\end{align}

\noindent \textbf{Recurrent loss:} We have so far considered the setting when static data is available. Instead, assume that we have access to unpaired but {\em ordered streams} $(x_1,x_2,\ldots,x_t,\ldots)$ and $(y_1,y_2\ldots, y_s,\ldots)$. Our motivating application is learning a mapping between two videos from different domains. One option is to ignore the stream indices, and treat the data as an unpaired {\em and unordered} collection of samples from $X$ and $Y$ (e.g., learn mappings between shuffled video frames). We demonstrate that much better mapping can be learnt by taking advantage of the temporal ordering. To describe our approach, we first introduce a recurrent temporal predictor $P_X$ that is trained to predict future samples in a stream given its past:

\begin{align}
 L_\tau(P_X) = \sum_t ||x_{t+1} - P_X(x_{1:t})||^2 ,
\end{align}

\noindent where we write $x_{1:t} = (x_1 \ldots x_t)$. 

\noindent \textbf{Recycle loss:} We use this temporal prediction model to define a new cycle loss across domains and {\em time} (Fig.~\ref{fig:overview}-c) which we refer as a recycle loss:
\begin{align}
L_{r}(G_X, G_Y, P_Y) =  \sum_t ||x_{t+1} - G_X(P_Y(G_Y(x_{1:t})))||^2 ,
\label{eq:cycle2}
\end{align}

\noindent where $G_Y(x_{1:t}) = (G_Y(x_1), \ldots, G_Y(x_t))$.  Intuitively, the above loss requires {\em sequences} of frames to map back to themselves. We demonstrate that this is a much richer constraint when learning from unpaired data streams in Figure~\ref{fig:tasks}.

\noindent \textbf{Recycle-GAN:} We now combine the recurrent loss, recycle loss, and adversarial loss into our final Recycle-GAN formulation:

\begin{align}
&\min_{G,P} \max_{D} L_{rg}(G,P,D) =  L_{g}(G_X,D_X) + L_{g}(G_Y,D_Y) + \nonumber\\
&\lambda_{rx} L_{r}(G_X, G_Y, P_Y) + \lambda_{ry} L_{r}(G_Y, G_X, P_X) + \lambda_{{\tau}x} L_{\tau}(P_{X}) + \lambda_{{\tau}y} L_{\tau}(P_{Y}) . \nonumber
\end{align}

\noindent\textbf{Inference: } At test time, given an input video with frames $\{x_t\}$, we would like to generate an output video. The simplest strategy would be directly using the trained $G_Y$ to generate a video frame-by-frame $y_t = G_Y(x_t)$.  Alternatively, one could use the temporal predictor $P_Y$ to smooth the output:
\begin{align}
y_{t} = \frac{G_{Y}(x_{t}) + P_{Y}(G_{Y}(x_{1:t-1}))}{2} ,\nonumber
\end{align}

\noindent where the linear combination could be replaced with a nonlinear function, possibly learned with the original objective function. However, for simplicity, we produce an output video by simple single-frame generation. This allows our framework to be applied to both videos and single images at test-time, and produces fairer comparison to spatial approach.

\noindent\textbf{Implementation Details: } We adopt much of the training details from Cycle-GAN~\cite{ZhuPIE17} to train our spatial translation model, and Pix2Pix~\cite{pix2pix2016} for our temporal prediction model. The generative network consists of two convolution (downsampling with stride-2), six residual blocks, and finally two upsampling convolution (each with a stride 0.5). We use the same network architecture for $G_X$, and $G_Y$. The resolution of the images for all the experiments is set to $256 \times 256$. The discriminator network is a $70 \times 70$ PatchGAN~\cite{pix2pix2016,ZhuPIE17} that is used to classify a $70 \times 70$ image patch if it is real or fake. We set all $\lambda_s = 10$. To implement our temporal predictors $P_X$ and $P_Y$, we concatenate the last two frames as input to a network whose architecture is identical to U-Net architecture~\cite{pix2pix2016,RonnebergerFB15}.

%% file: experiments.tex
\section{Experiments}
\label{experiments}

\begin{figure*}[t]
\centering
\includegraphics[width=1\linewidth]{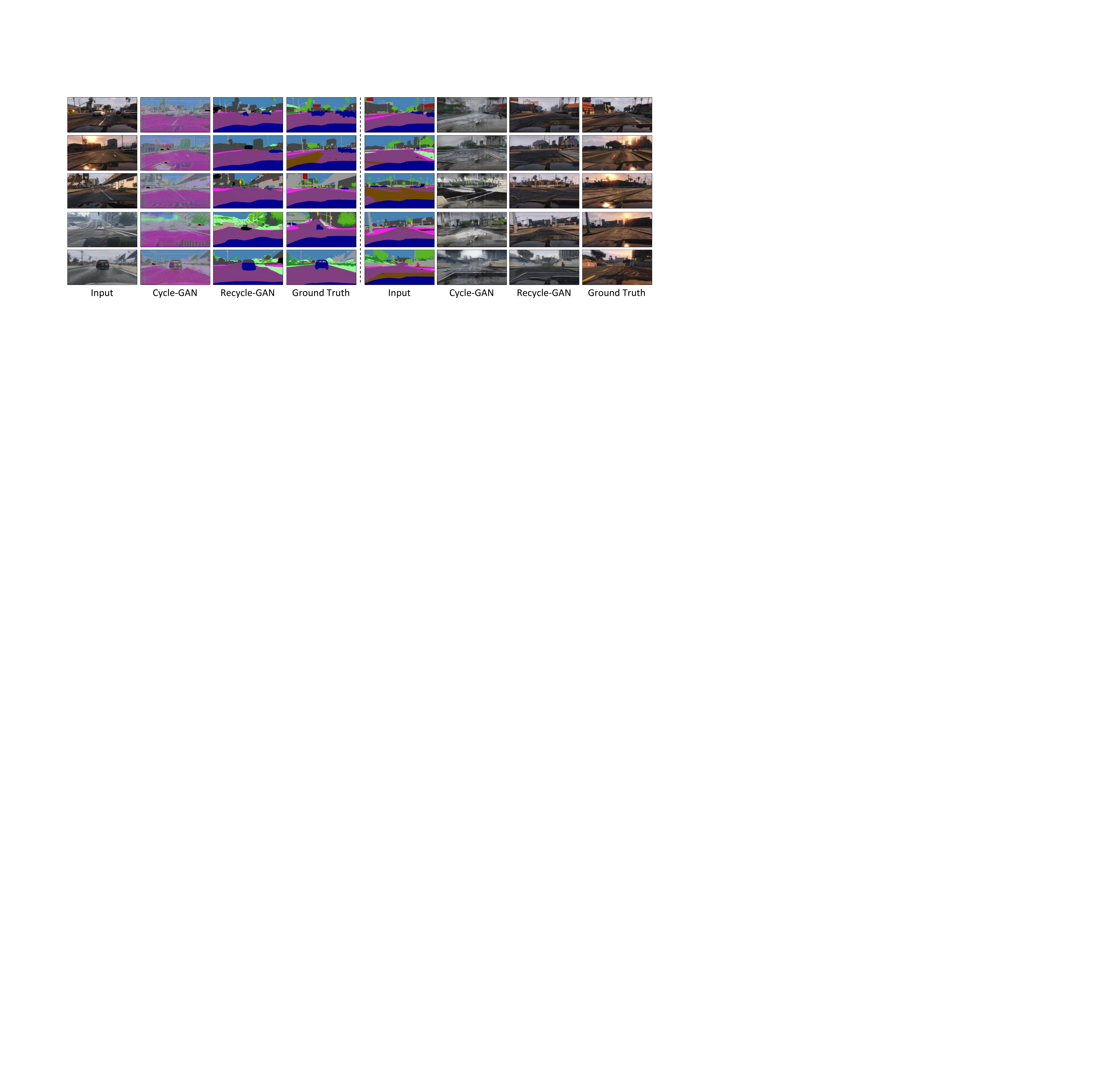}
\caption{ We compare the performance of our approach for image2labels and labels2image with Cycle-GAN~\cite{ZhuPIE17} on a held out data of Viper dataset~\cite{Richter_2017} for various environmental conditions.}
\label{fig:tasks}
\end{figure*}

We now study the influence of spatiotemporal constraints over spatial cyclic constraints. Because our key technical contribution is the introduction of temporal constraints in learning unpaired image mappings, the natural baseline is Cycle-GAN~\cite{ZhuPIE17}, a widely adopted approach for exploiting spatial cyclic consistency alone for an unpaired image translation. We first present quantitative results on domains where ground-truth correspondence between input and output videos are known (e.g., a video where each frame is paired with a semantic label map). Importantly, this correspondence pairing is {\em not available} to either Cycle-GAN or Recycle-GAN, but used only for evaluation. We then present qualitative results on a diverse set of videos with unknown correspondences, including video translations across different human faces and temporally-intricate events found in nature (flowers blooming, sunrise/sunset, time-lapsed weather progressions).

\subsection{Quantitative Analysis}
\label{spatiotemporal}

We use publicly available Viper~\cite{Richter_2017} dataset  for image-to-labels and labels-to-image to evaluate our findings. This dataset is collected using computer games with varying realistic content and provides densely annotated pixel-level labels. Out of the 77 different video sequences consisting of diverse environmental conditions, we use 57 sequences for training our model and baseline. The held-out 20 sequences are used for evaluation. The goal for this evaluation is not to achieve the state-of-the-art performance but to compare and understand the advantage of spatiotemporal cyclic consistency over the spatial cyclic consistency~\cite{ZhuPIE17}.

\begin{table}
\small{
\setlength{\tabcolsep}{5pt}
\def\arraystretch{1.2}
\center
\begin{tabular}{@{}l l c c c c c c }
\toprule
Criterion& Approach  & day  &   sunset & rain &  snow & night &  all \\
\midrule
\textbf{MP}& Cycle-GAN		      &	  35.8  &	38.9	 &  51.2	 &   31.8   &	27.4    & 	35.5	\\  	
& Recycle-GAN 	 		      &	  \textbf{48.7}    &	\textbf{71.0}	 &  \textbf{60.9} 	 &   57.1    &	\textbf{45.2}    & 	\textbf{56.0}	\\
& Combined				      &	  \textbf{48.7}   &	70.0	 & 60.1 	 &   \textbf{58.9}   &	33.7    & 	53.7	\\
\midrule
\textbf{AC} & Cycle-GAN		      &	  7.8  &	6.7	 &  	7.4 &    7.0  &	  4.7  & 	7.1 \\  	
& Recycle-GAN 	  		      &	  11.9  & 12.2	 &   \textbf{10.5}	 &    11.1   &	 \textbf{6.5}   & 	11.3	\\
& Combined				      &	  \textbf{12.6}   &	\textbf{13.2}	 & 10.1 	 &   \textbf{13.3}  &	5.9    & 	\textbf{12.4}	\\
\midrule
\textbf{IoU} & Cycle-GAN		      &	 4.9   &	3.9	 &  	4.9  &    4.0  &	  2.2  & 	4.2	\\  	
& Recycle-GAN 	 		      &	  7.9     &	9.6	 &   7.1	 &    8.2   &	 \textbf{4.1}   & 8.2		\\
& Combined				      &	  \textbf{8.5}   &	 \textbf{13.2}	 & \textbf{10.1} 	 &   \textbf{9.6}   &	3.1    & 	\textbf{8.9}	\\
\bottomrule
\end{tabular}
\caption{\textbf{Image-to-Labels (Semantic Segmentation):} We use the Viper~\cite{Richter_2017} dataset to evaluate the performance improvement when using spatiotemporal constraints as opposed to only spatial cyclic consistency~\cite{ZhuPIE17}. We report results using three criteria: (1). Mean Pixel Accuracy (\textbf{MP}); (2). Average Class Accuracy (\textbf{AC}); and (3). Intersection over union (\textbf{IoU}). We observe that our approach achieves better performance than prior work, and combining both leads to further better performance.}
\label{tab:seg_viper}}
\end{table}

While the prior works~\cite{pix2pix2016,ZhuPIE17} have mostly used Cityscapes dataset~\cite{Cordts2016Cityscapes}, we could not use it for our evaluation. Primarily the labelled images in Cityscapes are not continuous video sequences. That is consecutive frame pairs are quite different from other another. As such it is not trivial to use a temporal predictor. We used Viper as a proxy for Cityscapes because the task is similar and that dataset contains dense video annotations. Additionally, a concurrent work~\cite{Bashkirova2018} on unsupervised video-to-video translation also uses Viper dataset for evaluation. However, their experiments are restricted to a small subset of sequences from daylight category. In this work, we use all the environmental conditions available in the dataset.

\noindent\textbf{Image-to-Labels :}  In this setting, we input a RGB image to the generator that output segmentation label maps. We compute three metrics to compare the output ofs two approaches: (1). Mean Pixel Accuracy (\textbf{MP}); (2). Average Class Accuracy (\textbf{AC}); (3). Intersection over Union (\textbf{IoU)}. These statistics are computed using the ground truth for the held-out sequences from different environmental conditions. Table~\ref{tab:seg_viper} contrast the performance of our approach (Recycle-GAN) with Cycle-GAN. We observe that Recycle-GAN achieves better performance than Cycle-GAN, and combining both losses further improved it. 

\noindent\textbf{Labels-to-Image :} In this setting, we input a segmentation label map to generator and output an image that is close to a real image. The goal of this evaluation is to compare the quality of output images obtained from both approaches. We follow Pix2Pix~\cite{pix2pix2016} for this evaluation.  We use the generated images from each of the algorithm with a pre-trained FCN model. We then compute the performance of synthesized images against the real images to compute a normalized FCN-score. Higher performance on this criterion suggest that generated image is closer to the real images. Table~\ref{tab:seg2img_viper} compares the performance of our approach with Cycle-GAN. We observe that Recycle-loss does better than Cycle-loss, and combining both the losses led to significantly better outputs. Figure~\ref{fig:tasks} qualitatively compares our approach with Cycle-GAN.

\begin{table}
\small{
\setlength{\tabcolsep}{3pt}
\def\arraystretch{1.2}
\center
\begin{tabular}{@{}l c c c c c c }
\toprule
Approach & day  &   sunset & rain &  snow & night &  all \\
\midrule
Cycle-GAN   	&	0.33    		   &	0.27	 		&  0.39 		 &  0.29   		   &	0.37    		& 	0.30	\\  	
Recycle-GAN   &	0.33  	 	   &	0.51	 		&  0.37 		 &  0.43   		   &  0.40			& 	0.39 \\
Combined         &	 \textbf{0.42}         &	\textbf{0.61}	 & \textbf{0.45} 	 &  \textbf{0.54}   &  \textbf{0.49}   	& 	\textbf{0.48}	\\
\bottomrule
\end{tabular}
\caption{\textbf{Normalized FCN score for Labels-to-Image: } We use a pre-trained FCN-style model to evaluate the quality of synthesized images over real images using the Viper~\cite{Richter_2017} dataset. Higher performance on this criteria suggest that the output of a particular approach produces images that look closer to the real images.}
\label{tab:seg2img_viper}}
\end{table}

In these experiments, we make two observations: (i) Cycle-GAN learns a good translation model within a few initial iterations (seeing only a few examples) but this model degraded as reconstruction loss started to decrease. We believe that minimizing reconstruction loss alone on input led it to a bad local minima. Our formulation provided more constraints and led it to a better local minima; (ii) Cycle-GAN learns a better translation model for Cityscapes as opposed to Viper. Cityscapes consists of images from mostly daylight and agreeable weather. This is not the case with Viper as it is rendered. It has a large and varied distribution of different illumination (day, night) and weather conditions (snow, rain). This makes it harder to learn a good mapping because for each labelled input as there are potentially many output images. We find that standard conditional GANs suffer from mode collapse in such scenarios and produce ``average'' outputs (as pointed by prior works~\cite{pixelnn}). Our experiments suggest that having more constraints help ameliorate such challenging translation problems. 

\begin{figure*}[t]
\centering
\includegraphics[width=1\linewidth]{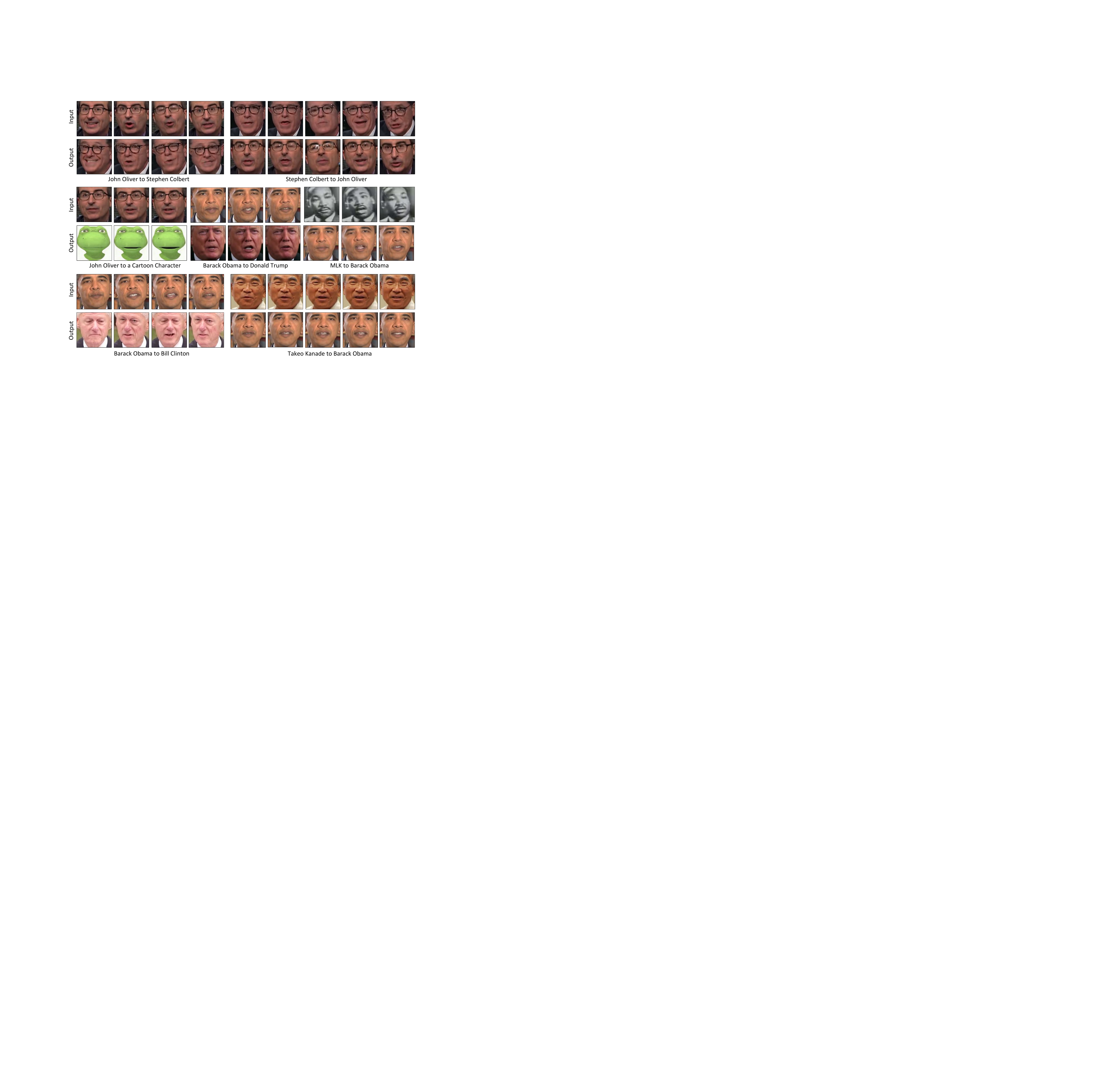}
\caption{\textbf{Face to Face:} The top row shows multiple examples of face-to-face between John Oliver and Stephen Colbert using our approach. The bottom row shows example of translation from John Oliver to a cartoon character, Barack Obama to Donald Trump, and  Martin Luther King Jr. (MLK) to Barack Obama. Without any input alignment or manual supervision, our approach could capture stylistic expressions for these public figures. As an example, John Oliver's dimple while smiling, the shape of mouth characteristic of Donald Trump, and the facial mouth lines and smile of Stephen Colbert. More results and videos are available on our project webpage.}
\label{fig:face_to_face}
\end{figure*}

\subsection{Qualitative Analysis}

\noindent\textbf{Face to Face: }  We use publicly available videos of various public figures for face to face translation. The faces are extracted using facial keypoints generated using the OpenPose Library\cite{cao2017realtime} and are manually curated to remove false positives. Figure~\ref{fig:face_to_face} shows an example of face-to-face translation between John Oliver and Stephen Colbert, Barack Obama to Donald Trump, and Martin Luther King Jr. (MLK) to Barack Obama, and John Oliver to a cartoon character. Note that without any supervisory signal or manual alignment, our approach learns face-to-face translation and captures stylistic expression for these personalities, such as the dimple on the face of John Oliver while smiling, the characteristic shape of mouth of Donald Trump, and the mouth lines for Stephen Colbert.  

\begin{figure*}[t]
\centering
\includegraphics[width=1\linewidth]{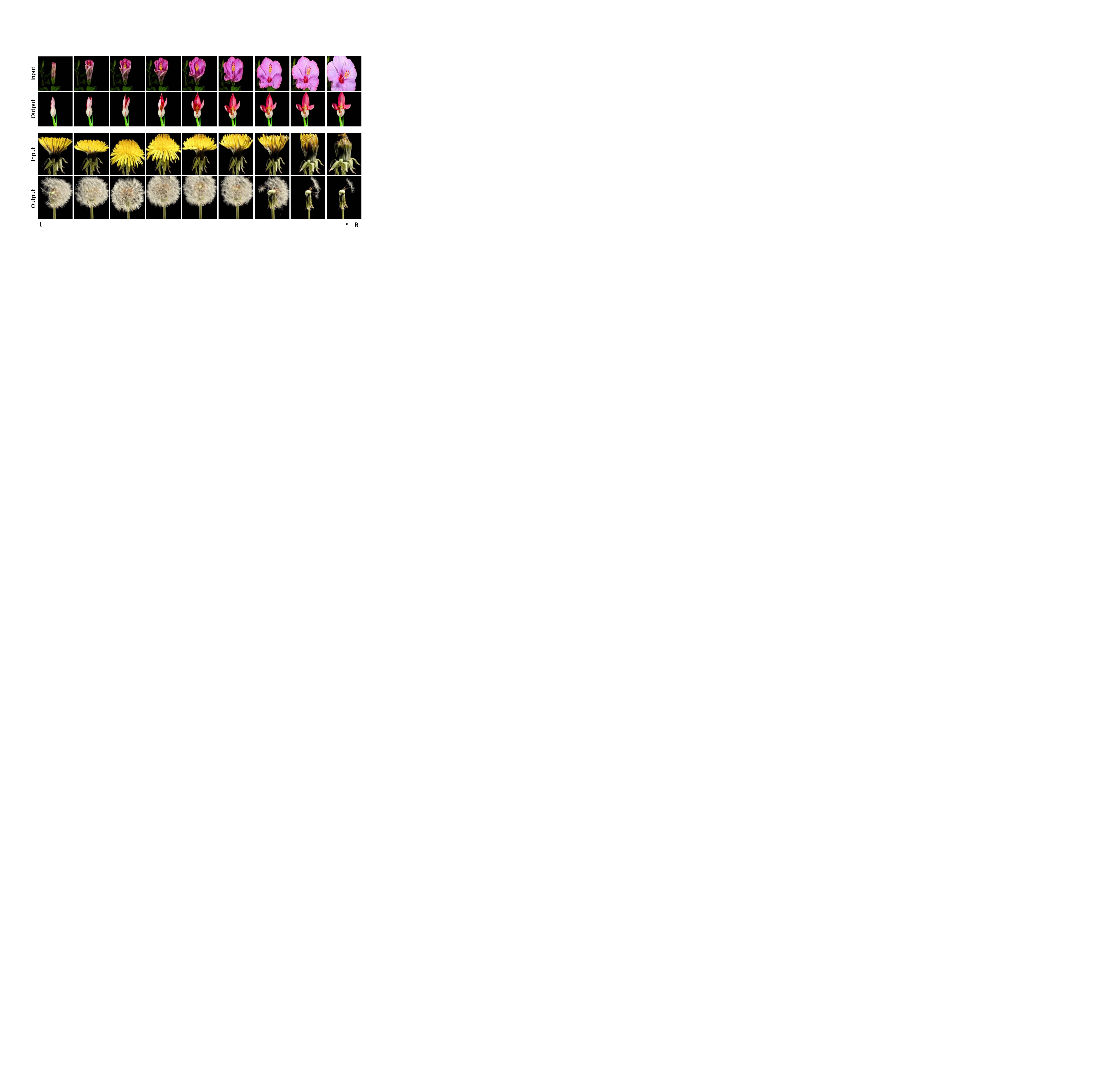}
\caption{\textbf{Flower to Flower:} We shows two examples of flower-to-flower translation. Note the smooth transition from Left to Right. These results can be best visualized using videos on our project webpage.}
\label{fig:flowers_to_flowers}
\end{figure*}

\noindent\textbf{Flower to Flower: } Extending from faces and other traditional translations, we demonstrate our approach for flowers. We extracted the time-lapse of various flowers from publicly available videos. The time-lapses show the blooming of different flowers but without any sync. We use our approach to align the content, i.e. both flowers bloom or die together. Figure~\ref{fig:flowers_to_flowers} shows the output of approach to learn association between the events of different flowers' life.

\subsection{Video Manipulation via Retargeting}

\noindent\textbf{Clouds \& Wind Synthesis: } Our approach can be used to synthesize a new video that has the required weather condition such as clouds and wind  without the need for recapturing. We use the given video, and video data from target environmental condition as two domains in our experiment. The conditional video, and trained translation model is then used to generate the required output. 

\begin{figure*}[t]
\centering
\includegraphics[width=1\linewidth]{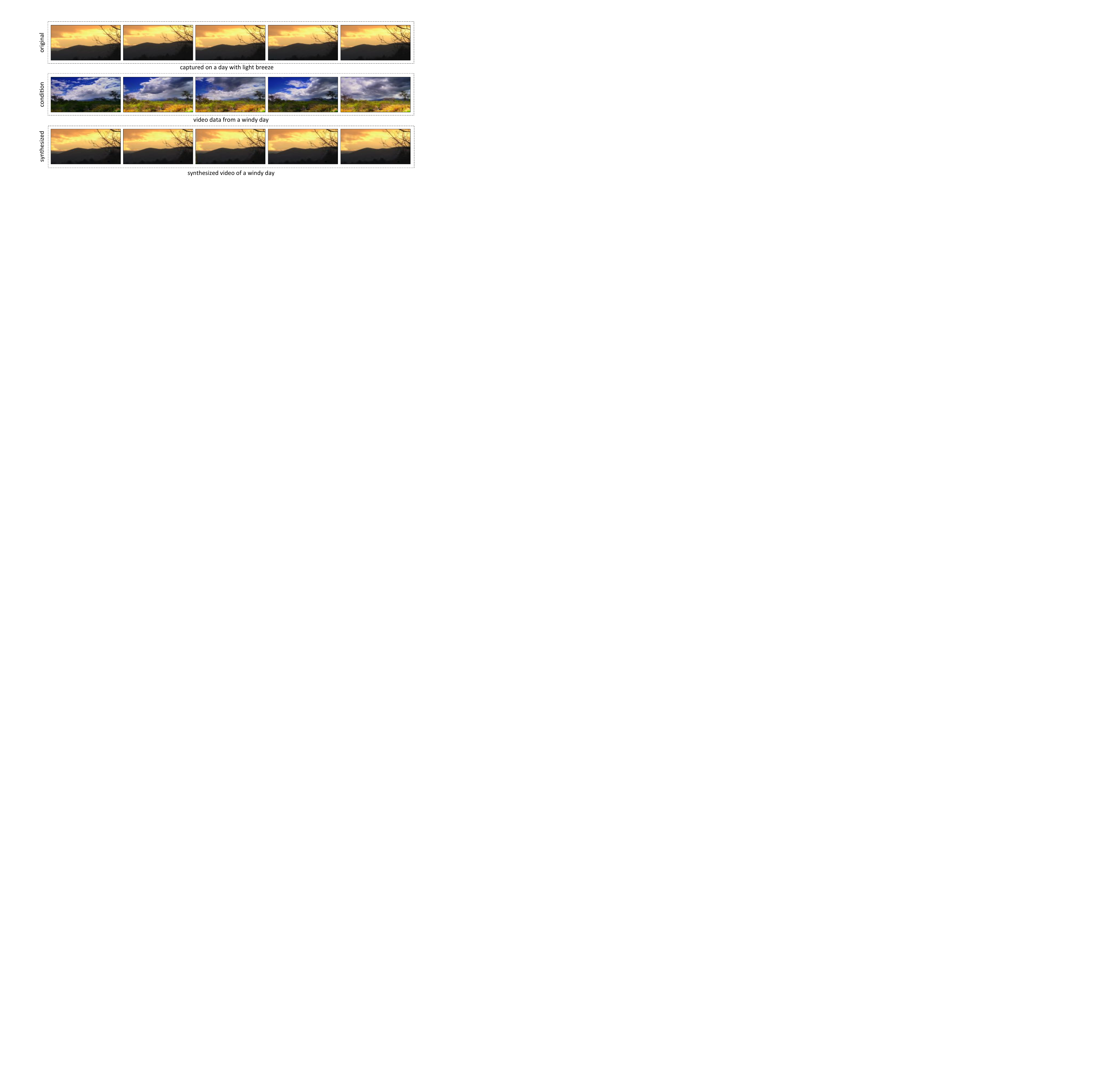}
\caption{\textbf{Synthesizing Clouds \& Winds:} We use our approach to synthesize clouds and winds. The top row shows example frames of a video captured on a day with light breeze. We condition it on video data from a windy data (shown in second row) by learning a transformation between two domains using our approach. The last row shows the output synthesized video with the clouds and trees moving faster (giving a notion of wind blowing). Refer to the videos on our project webpage for better visualization and more examples.}
\label{fig:wind_synthesis}
\end{figure*}

We collected the video data for various wind and cloud conditions, such as calm day or windy day for this experiment. We convert a calm-day to a windy-day, and a windy-day to a calm-day using our approach, without modifying the aesthetics of the place. Figure~\ref{fig:wind_synthesis} shows an example of synthesizing clouds and winds on a windy day at a place when the only information available was a video captured at same place with a light breeze.  Additional videos for these clouds and wind synthesis are available on our project webpage.

\begin{figure*}[t]
\centering
\includegraphics[width=1\linewidth]{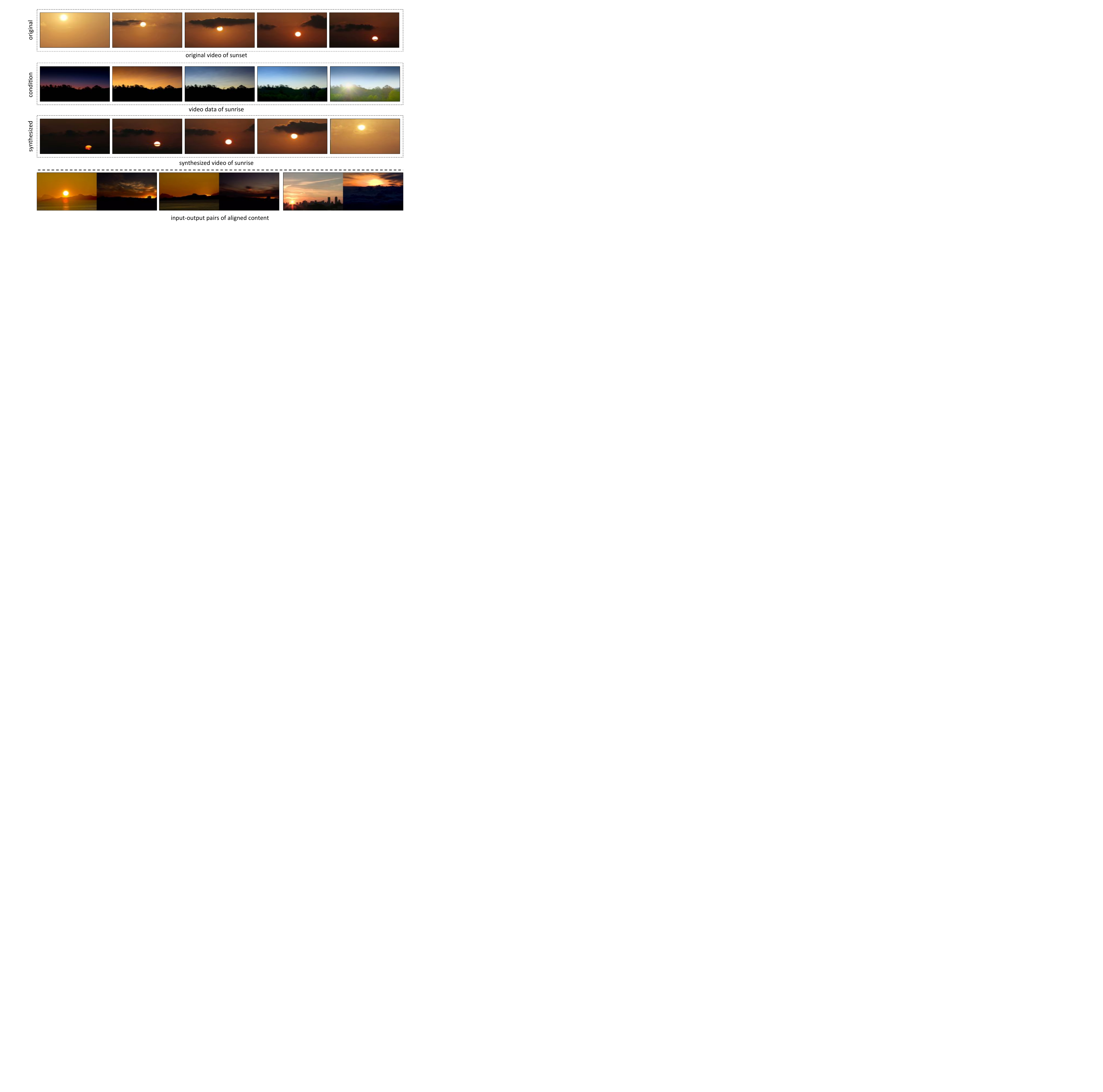}
\caption{\textbf{Sunrise \& Sunset:} We use our approach to manipulate and align the videos of sunrise and sunset. The top row shows example frames from a sunset video. We condition it on video data of sunrise (shown in second row) by learning a transformation between two domains using our approach. The third row shows example frames of new synthesized video of sunrise. Finally, the last row shows random examples of input-output pair from different sunrise and sunset videos. Videos and more examples are available on our project webpage.}
\label{fig:sunrise_sunset}
\end{figure*}

\noindent\textbf{Sunrise \& Sunset: } We extracted the sunrise and sunset data from various web videos, and show how our approach could be used for both video manipulation and content alignment. This is similar to our experiments on clouds and wind synthesis. Figure~\ref{fig:sunrise_sunset} shows an example of synthesizing a sunrise video from an original sunset video by conditioning it on a random sunrise video. We also show examples of alignment of various sunrise and sunset scenes. 

\noindent\textbf{Note:} We refer the reader to our project webpage for different videos synthesized using our approach. We have also added the results using both Cycle-loss and Recycle-loss on our project webpage.

\subsection{Human Studies}

We performed human studies on the synthesized output, particularly faces and flowers, following the protocol of MoCoGAN~\cite{tulyakov2017mocogan} who also evaluated videos. However, our analysis consist of three parts: (1). We show the synthesized videos individually from both Cycle-GAN and ours  to $15$ sequestered human subjects, and asked them if it is a real video or a generated video. The subjects misclassified $28.3\%$ times  generated videos from our approach as real, and $7.3\%$ times for Cycle-GAN. (2). We show the synthesized videos from Cycle-GAN and our approach simultaneously, and asked them to tell which one looks more natural and realistic. Human subjects chose the videos synthesized from our approach $76\%$ times, $8\%$ times Cycle-GAN, and $16\%$ times they were confused. (3). We show the video-to-video translation. This is an extension of (2), except now we also include input and ask which translation looks more realistic and natural. We showed each video to $15$ human subjects. The human subjects selected our approach $74.7\%$ times, $13.3\%$ times they selected Cycle-GAN, and $12\%$ times they were confused. From the human study, we can clearly see that combining spatial and temporal constraints lead to better retargeting.

\subsection{Failure Example: Learning association beyond data distribution} 

We show an example of transformation from a real bird to a origami bird to demonstrate a case where our approach failed to learn the association. The real bird data was extracted using web videos, and we used the origami bird from the synthesis of Kholgade et al.~\cite{OM3D2014}. Figure~\ref{fig:failures} shows the synthesis of origami bird conditioned on the real bird. While the real bird is sitting, the origami bird stays and attempts to imitate the actions of real bird. The problem comes when the bird begins to fly.  The initial frames when the bird starts to fly are fine. After some time the origami bird reappears. From an association perspective, the origami bird should not have reappeared. Looking back at the training data, we found that the original origami bird data does not have a example of frame without the origami bird, and therefore our approach is not able to associate an example when the real bird is no longer visible. Perhaps, our approach could only learn to interpolate over a given data distribution and fails to capture anything beyond it. A possible way to address this problem is by using a lot of training data so that it encapsulates all possible scenarios for an effective interpolation.

%% file: discussion.tex
\section{Discussion \& Future Work}

\begin{figure*}[t]
\centering
\includegraphics[width=1\linewidth]{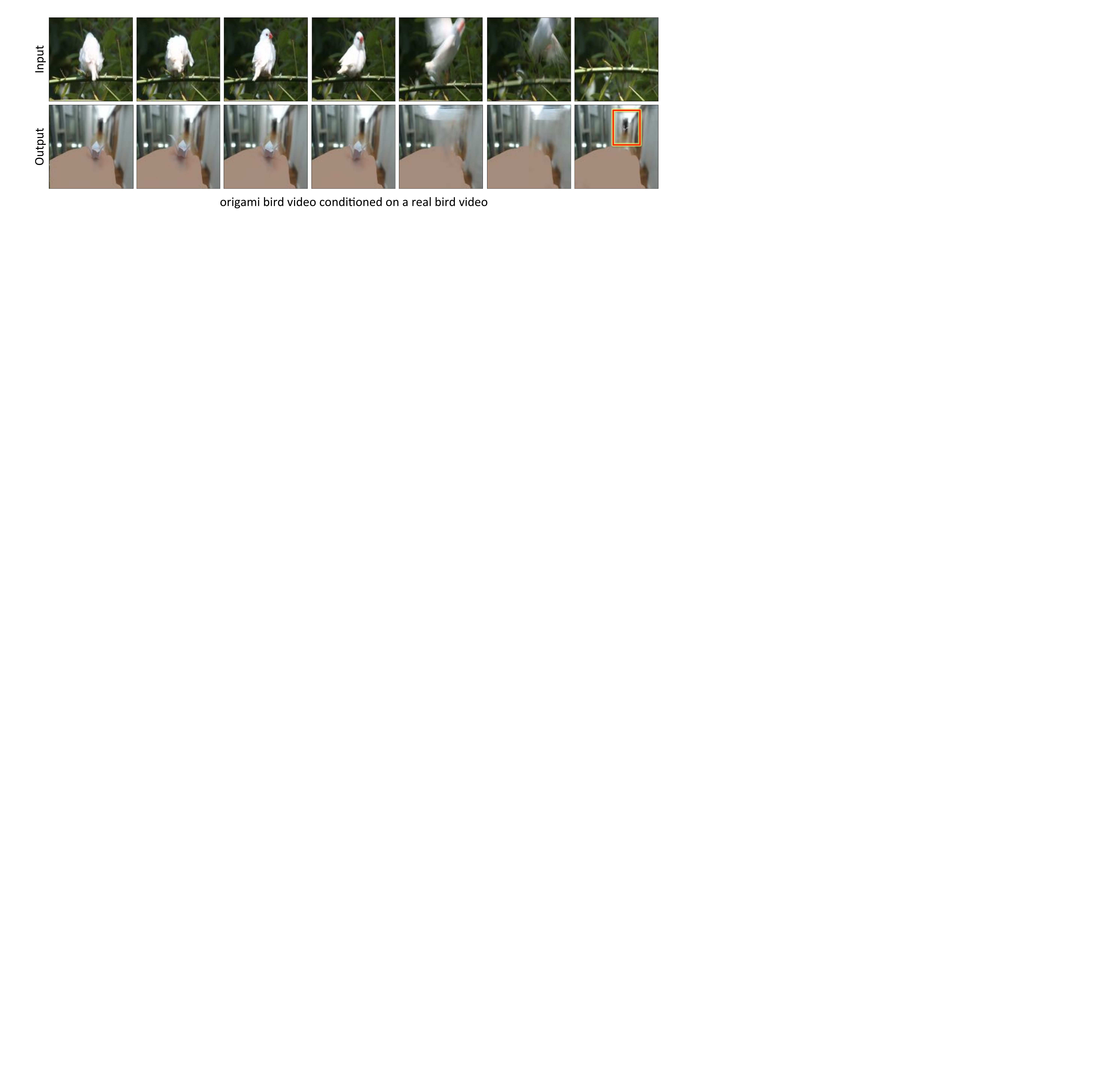}
\caption{\textbf{Failure Example:} We present the failure in association/synthesis for our approach using a transformation from a \emph{real} bird to an \emph{origami} bird. While the origami bird (output) is trying to imitate the real bird (input) when it is sitting (Column 1 - 4), and also flies away when the real bird flies (Column 5 - 6). We observe that it reappears after sometime (red bounding box in Column 7) in a flying mode while the real bird didn't exist in the input. Our algorithm is not able to make transition of association when the real bird is completely invisible, and so it generated a random flying origami.}
\label{fig:failures}
\end{figure*}

In this work, we explore the influence of spatiotemporal constraints in learning video retargeting and image translation. Unpaired video/image translation is a challenging task because it is unsupervised, and lacks any correspondences between training samples from the input and output space. We point out that many natural visual signals are inherently spatiotemporal in nature, which provides strong temporal constraints for free. This results in significantly better mappings. We also point out that unpaired and unsupervised video retargeting and image translation is an under-constrained problem. More constraints using auxiliary tasks from the visual data itself (as used for  other vision tasks~\cite{Meister:2018:UUL,zhou2017unsupervised}) could help in learning better transformation models.

Recycle-GANs learn both a mapping function and a recurrent temporal predictor. Thus far, our results make use of only the mapping function, so as to facilitate fair comparisons with previous work. But it is natural to synthesize target videos by making use of both the single-image translation model and the temporal predictor. Additionally, the notion of style in video retargeting can be incorporated more precisely by using spatiotemporal generative models as this would allow to even learn the speed of generated output. E.g. Two people may have different ways of content delivery and that one person can take longer than other to say the same thing. A true notion of style should be able to generate even this variation in time required for delivering speech/content. We believe that better spatiotemporal neural network architecture could attempt this problem in near future. Finally, our work could also utilize the concurrent approach from Huang et al.~\cite{huang2018munit} to learn a one-to-many translation model.